\documentclass{article}

\PassOptionsToPackage{numbers,compress}{natbib}
\usepackage[final]{neurips_2024}

\usepackage{amsmath,amsfonts,bm}

\def\eqref#1{equation~\ref{#1}}

\def\1{\bm{1}}

\def\vc{{\bm{c}}}

\def\vx{{\bm{x}}}

\def\mI{{\bm{I}}}

\def\mQ{{\bm{Q}}}

\DeclareMathAlphabet{\mathsfit}{\encodingdefault}{\sfdefault}{m}{sl}
\SetMathAlphabet{\mathsfit}{bold}{\encodingdefault}{\sfdefault}{bx}{n}

\newcommand{\E}{\mathbb{E}}

\usepackage[utf8]{inputenc} 
\usepackage[T1]{fontenc}    
\usepackage{url}            
\usepackage{booktabs}       
\usepackage{amsfonts}       
\usepackage{nicefrac}       
\usepackage{microtype}      
\usepackage{xcolor}         

\usepackage{amsmath, array}
\usepackage{amssymb}
\usepackage{bm}
\usepackage{textcomp}
\usepackage{siunitx}
\usepackage{tikz}
\usepackage{graphicx}
\usetikzlibrary{positioning, shapes.geometric, arrows}
\usetikzlibrary{decorations.pathreplacing}
\usetikzlibrary{calc}
\usepackage{comment}
\usepackage{subcaption}

\usepackage[colorlinks, allcolors=magenta]{hyperref}
\usepackage[capitalise]{cleveref}
\usepackage{threeparttable}  

\usepackage[normalem]{ulem}
\usepackage{color}

\graphicspath{{figs/}}

\title{DiffBatt: A Diffusion Model for Battery Degradation Prediction and Synthesis}

\author{%
  Hamidreza Eivazi$^{1,5,}$\thanks{Corresponding author: \texttt{he76@tu-clausthal.de}~(https://orcid.org/0000-0003-3650-4107)} \quad
  André Hebenbrock$^{2,5}$ \quad
  Raphael Ginster$^{3,5}$ \quad
  Steffen Blömeke$^{4,5}$ \\
  \textbf{Stefan Wittek}$^1$ \quad
  \textbf{Christoph Herrmann}$^{4,5}$ \quad
  \textbf{Thomas S. Spengler}$^{3,5}$ \\
  \textbf{Thomas Turek}$^{2,5}$ \quad
  \textbf{Andreas Rausch}$^{1,5}$\\ \\
  $^1$\,Institute for Software and Systems Engineering, \\Clausthal University of Technology, Germany
  \\[2pt]$^2$\,Institute of Chemical and Electrochemical Process Engineering, \\Clausthal University of Technology, Germany
  \\[2pt]$^3$\,Institute of Automotive Management and Industrial Production, \\Technische Universität Braunschweig, Germany
  \\[2pt]$^4$\,Institute of Machine Tools and Production Technology, \\Technische Universität Braunschweig, Germany
  \\[2pt]$^5$\,Battery LabFactory Braunschweig (BLB), \\Technische Universität Braunschweig, Germany
}

\begin{document}
\hypersetup{
  pdftitle={},
  pdfsubject={},
  urlcolor=magenta,
  citecolor=blue,
  linkcolor=red
  }

\maketitle

\begin{abstract}
Battery degradation remains a critical challenge in the pursuit of green technologies and sustainable energy solutions. Despite significant research efforts, predicting battery capacity loss accurately remains a formidable task due to its complex nature, influenced by both aging and cycling behaviors. To address this challenge, we introduce a novel general-purpose model for battery degradation prediction and synthesis, DiffBatt. Leveraging an innovative combination of conditional and unconditional diffusion models with classifier-free guidance and transformer architecture, DiffBatt achieves high expressivity and scalability. DiffBatt operates as a probabilistic model to capture uncertainty in aging behaviors and a generative model to simulate battery degradation. The performance of the model excels in prediction tasks while also enabling the generation of synthetic degradation curves, facilitating enhanced model training by data augmentation. In the remaining useful life prediction task, DiffBatt provides accurate results with a mean RMSE of 196 cycles across all datasets, outperforming all other models and demonstrating superior generalizability. This work represents an important step towards developing foundational models for battery degradation.
\end{abstract}

\section{Introduction}
\subsection{Lithium-ion batteries}
Lithium-ion (Li-ion) batteries are key technologies in the field of energy storage, with applications spanning portable electronics and electric vehicles \cite{Blomgren2017TheBatteries}. The prominence of these batteries is largely attributable to their high energy density, which enables substantial energy storage within a compact and lightweight form factor. Moreover, Li-ion batteries demonstrate an extended cycle life compared to other battery technologies, quantified in terms of charge and discharge cycles, thereby enhancing their cost-effectiveness for long-term usage. The low self-discharge rate of Li-ion batteries ensures minimal energy loss during periods of inactivity, which is a significant advantage over other battery technologies \cite{Galeotti2015PerformanceSpectroscopy,Vetter2005AgeingBatteries}. Nevertheless, several challenges remain. Safety continues to be a major issue, as mechanical damage or improper handling can potentially lead to hazardous events such as thermal runaway \cite{Chombo2020ABattery}. Furthermore, the economic and environmental implications of Li-ion battery production, recycling, and disposal present additional complexities that warrant ongoing investigation \cite{Blomeke2022,Ginster2024CircularImpacts}. One persistent challenge that continues to impact their long-term performance and reliability is capacity degradation.

\subsection{Capacity degradation}
The phenomenon of capacity degradation in Li-ion batteries is a multifaceted issue that encompasses both the effects of aging and the effects of cycling. Aging behavior, which is often referred to as calendar aging, pertains to the decline in battery performance over time, irrespective of active usage. Factors such as ambient temperature, state of charge, and storage conditions play a significant role in this degradation mode. In contrast, cycling behavior, also termed cycle aging, is linked to the deterioration that batteries experience during charge and discharge cycles. High charge-discharge rates and frequent cycling result in the accumulation of irreversible changes within the battery's electrochemical structure. This degradation is driven by a number of factors, including the formation of a solid electrolyte interphase layer, electrolyte decomposition, and the growth of lithium plating \cite{Broussely2005MainBatteries,Edge2021LithiumKnow}. Both aging and cycling behaviors collectively result in overall degradation, reducing the battery's capability to store and deliver electric charge \cite{Rubenbauer2017DefinitionsGrids} over its operational lifespan. In addition, the aging of batteries is very individual depending on, e.g., usage behavior and environmental conditions, which makes a basic understanding difficult and hinders individual battery management. Despite extensive research, accurately predicting the rate and extent of capacity loss remains a formidable challenge \cite{OKane2022Lithium-ionIt}. Therefore, advanced modeling techniques, including machine learning, are increasingly being employed to provide more accurate predictions of the battery degradation processes.

\subsection{Machine learning in battery life prediction}
The degradation of a battery can be quantified by using key performance indicators, including the state of health (SOH) and the remaining useful life (RUL) \cite{Li2022TowardsReview}. The SOH is a measure of the current capacity of a battery relative to its original capacity, expressed as a percentage, and provides insight into the extent of capacity degradation \cite{Cui2022ANetwork,Ren2023ABatteries}. In contrast, the RUL is a predictive measure that estimates the remaining operational cycles of a battery before it reaches defined performance criteria \cite{Li2023TheBatteries}.

Methods for battery degradation modeling can be classified according to \citet{Rauf2022MachineModelling} into four domains: i) physics-based models, ii) empirical models, iii) data-driven methods (DDMs), and iv) hybrid methods. Among these various domains, DDMs are emerging as a prominent technique for developing battery degradation models. This is due to the flexibility and independence from specific model assumptions that these approaches offer. In the domain of DDMs, machine learning (ML) methods are widely regarded as one of the most effective approaches for estimating RUL and SOH, due to their ability to address non-linear problems \cite{Rauf2022MachineModelling}. Since all battery RUL and SOH prediction tasks are effectively regression problems, supervised learning is the most commonly used approach in ML battery studies. 

Recent literature reviews \cite{Li2022TowardsReview,Li2023TheBatteries, Rauf2022MachineModelling,Ren2023ABatteries,Wang2021} indicate that various ML methods are  utilized for modeling battery degradation. In the area of artificial neural networks, shallow neural networks can capture nonlinear relationships among an arbitrary number of inputs and outputs, however, they are hindered by slow training processes and a propensity to converge at local minima \cite{Li2022TowardsReview,Ren2023ABatteries}. In contrast, deep learning algorithms demonstrate superior performance in managing large datasets due to their specialized architectures. They provide higher accuracy and enhanced generalization capabilities but incur significant computational costs \cite{Ren2023ABatteries}. Techniques such as convolutional neural networks (CNNs), recurrent neural networks (RNNs), and long short-term memory (LSTM) networks are commonly employed in this context \cite{Li2023TheBatteries,Rauf2022MachineModelling,Wang2021}. Additionally, support vector machines achieve a commendable balance between generalization capability and estimation accuracy. However, they may struggle with scalability on larger datasets \cite{Ren2023ABatteries}. Similarly, relevance vector machines have the disadvantage of requiring extensive datasets, which results in significant computational complexity. However, they offer the advantage of high accuracy, robust learning capabilities, and the capacity to generate predictions with associated probability distributions \cite{Rauf2022MachineModelling}. Lastly, Gaussian process regression (GPR) methods are advantageous for their ability to quantify the uncertainty of estimated values, which is particularly valuable in practical applications. Nonetheless, GPR methods typically exhibit lower efficiency in high-dimensional spaces and can be computationally complex \cite{Li2022TowardsReview,Ren2023ABatteries}. Another recent approach, with a sole focus on the prediction of the SOH, is presented by \citet{Luo2023State-of-HealthLearning}, in which the authors introduce the methodology of diffusion models as a promising avenue for SOH prediction.

Despite the widespread application of ML models in battery degradation analysis, comparing these various approaches presents significant challenges. Many studies utilize different datasets, which are often not publicly available due to confidentiality concerns. To address this issue, BatteryML was developed by \citet{Zhang2023BatteryML:AnDegradation}, offering a standardized method for data representation that consolidates and harmonizes all accessible public battery datasets. Additionally, BatteryML establishes clear benchmarks for predicting RUL and includes a range of models, such as linear models, tree-based models, and neural networks, tailored for battery degradation prediction. Lastly, BatteryML also introduces the transformer architecture as a novel approach for predicting SOH and RUL \cite{Zhang2023BatteryML:AnDegradation}.

\subsection{Contribution to literature}
Our contribution represents a threefold advancement in the domain of battery degradation prediction. First, we introduce DiffBatt, a novel application of denoising diffusion probabilistic models (DDPMs) specifically tailored for estimating the SOH and RUL of Li-ion batteries. The presented approach is motivated by the need to handle the stochastic and intricate nature of battery degradation. This innovative approach employs the capabilities of diffusion models to more effectively capture the complex degradation behavior than traditional methods (\cref{subsec:architecture}). Secondly, our results demonstrate a significant improvement over existing benchmarks, showcasing superior accuracy and reliability in predictions (\cref{sec:results}). Finally, we establish the groundwork for further development by positioning our model as a foundation model, enabling future enhancements and adaptations to diverse energy storage related applications (\cref{sec:foundation_model}). All the codes and pre-trained models are available on GitHub:~\url{https://github.com/HamidrezaEiv/DiffBatt.git}.

\section{Methodology}

DDPMs \cite{ddpm} are an expressive and flexible family of generative models that utilize a parameterized Markov chain to produce high-quality samples that match the characteristics of the training data. The key idea behind DDPMs is to learn a reverse process, also known as the reverse diffusion or denoising process, which gradually removes noise from a sequence of noisy samples until it reveals the original signal. The forward process of a DDPM is a Markov chain that progressively adds noise to the data in the opposite direction of sampling. This process continues until the signal is completely obscured by noise. The model's objective is to learn a set of transformations that can effectively undo this noise accumulation and recover the original signal. 

The training of DDPMs involves using variational inference to optimize the model parameters, with the goal of minimizing the difference between the generated samples and the true data distribution. This is achieved by estimating the lower bound of the loss function along a large number of diffusion steps, which are computed iteratively during the training process. The resulting model can generate new, diverse samples that resemble the original data, making it useful for applications such as image and audio synthesis \cite{dhariwal2021diffusion,Ho2022,yang2024surveydiffusionmodelstime}, data augmentation \cite{luzi2024boomeranglocalsamplingimage}, and more \cite{bastek2024physicsinformeddiffusionmodels,Frrutter2024,Li2024,Zhang2024}.

We follow the principles proposed by \citet{diffusionguidance} for classifier-free diffusion guidance to increase sample quality while decreasing sample diversity in diffusion models. Classifier-free guidance serves to achieve similar objectives for performing truncated or low-temperature sampling in certain generative models, such as generative adversarial networks (GANs) and flow-based models. The intended outcome is a decrease in sample diversity accompanied by an increase in individual sample quality. Examples are truncation in BigGAN \citep{brock2019largescalegantraining} and low-temperature sampling in Glow \citep{Glow}, which lead to a trade-off curve between the Fréchet inception distance (FID) score and the inception score. This enables the flexibility to generate high-quality or more diverse samples when predicting or synthesizing battery degradation, respectively.

DiffBatt is based on a DDPM enhanced with transformer models and utilizes classifier-free diffusion guidance for conditional generative modeling. In the following, we provide a brief discussion on the methodological aspects of DDPMs and classifier-free guidance. 

\subsection{Background}

Denoising diffusion models learn to systematically transform a sample of a simple prior, typically a unit Gaussian, to a sample from an unknown data distribution $q(\vx)$. In a fixed \textit{forward process}, a given data sample $\vx_0 \sim q(\vx)$ is corrupted by Gaussian noise according to a variance schedule $\{\beta_t \in (0, 1)\}_{t=1}^T$ over the course of $T$ timesteps
\begin{equation}
  q(\vx_{1:T} \vert \vx_{0})=\prod_{t=1}^{T} q(\vx_{t} \vert \vx_{t-1}), \quad q(\vx_{t} \vert \vx_{t-1})=\mathcal{N}(\vx_{t} ; \sqrt{1-\beta_{t}} \vx_{t-1}, \beta_{t} \mI).
\end{equation}
The so-called \textit{reverse process} is considered to generate new samples
\begin{equation}
q(\vx_{0: T})=p(\vx_{T}) \prod_{t=1}^{T} q(\vx_{t-1} \vert \vx_{t}), \quad q(\vx_{t-1} \vert \vx_{t})=\mathcal{N}(\vx_{t-1} ; \bm{\mu}(\vx_{t}, t), \bm{\Sigma}(\vx_{t},t)),
\end{equation}
in which the unknown true inverse conditional distribution $q(\vx_{t-1}\vert\vx_t)$ is approximated by a neural network $p_{\theta}(\vx_{t-1} \vert \vx_{t})$ parameterized by $\theta$. The model seeks to learn an estimator for the mean parameter $\bm{\mu}_{\theta}(\vx_t, t)$, under the constraint that the covariance remains unchanged as
\begin{equation}
    \bm{\Sigma}\left(\vx_t, t\right)=\frac{1-\bar{\alpha}_{t-1}}{1-\bar{\alpha}_t} \beta_t \mI=\Sigma_t \mI
\end{equation}
with $\bar{\alpha}_{t}=\prod_{i=1}^{t} \alpha_{i}, \, \alpha_{t}=1-\beta_{t}$. The mean $\bm{\mu}_{\theta}(\vx_t, t)$ is parameterized as
\begin{equation}
  \label{eq:reparameterization_mu}
  \bm{\mu}_{\theta}(\vx_t,t)=\frac{1}{\sqrt{\alpha_{t}}}\left(\vx_{t}-\frac{\beta_{t}}{\sqrt{1-\bar{\alpha}_{t}}} \bm{\epsilon}_{\theta}(\vx_t, t)\right),
\end{equation}  
where $\vx_t = \sqrt{\bar\alpha_t} \vx_0 + \sqrt{1-\bar\alpha_t}\bm{\epsilon}_t$ for $\bm{\epsilon}_t \sim \mathcal{N}(\bm{0}, \mI)$, $\bm{\epsilon}_{\theta}$ is a function approximator intended to predict $\bm{\epsilon}_t$ from $\vx_t$ and $\bm{\epsilon}_{t}$ indicates Gaussian noise to diffuse $\vx_0$ to $\vx_t$. The reverse process is trained to approximate the joint distribution of the forward process by optimizing the evidence lower bound. With the parameterizations suggested in \cite{Ho2022} the loss simplifies to
\begin{equation}
  L_{\mathrm{simple}}(\theta):=\E_{t, \vx_0, \bm{\epsilon}}\left[\left\| \bm{\epsilon}_t - \bm{\epsilon}_\theta(\vx_t, t) \right\|^2\right], \label{eq:vb_term_langevin_eps}
\end{equation}
resembling denoising score matching over multiple noise scales. To summarize, we can train the reverse process to predict $\bm{\epsilon}_t$. To learn a conditional model $p_{\theta}(\vx_0 \vert \vc)$ the diffusion model is extended by incorporating the conditioning variable $\vc$ into the reverse process
\begin{equation}
\begin{split}
    & p_\theta(\vx_{0: T}\vert \vc)=p(\vx_{T}) \prod_{t=1}^{T} q(\vx_{t-1} \vert \vx_{t}, \vc), \\ & p_\theta(\vx_{t-1} \vert \vx_{t}, \vc)=\mathcal{N}(\vx_{t-1} ; \bm{\mu}_\theta(\vx_{t}, t,\vc), \bm{\Sigma}_\theta(\vx_{t},t,\vc)).
\end{split}
\end{equation}
Following \textit{classifier-free guidance} \cite{diffusionguidance}, we train an unconditional DDPM alongside the conditional one by randomly setting the conditioning information $\vc$ to a null token with probability $p_{\rm uncond}$, set as a hyperparameter. To generate samples, we combine the scores from the conditional and unconditional models 
\begin{equation}
  \tilde{\bm{\epsilon}}_\theta(\vx_t, t, \vc) = (1+w)\bm{\epsilon}_\theta(\vx_t, t, \vc) - w\bm{\epsilon}_{\theta}(\vx_t, t), \label{eq:classifier_free_score}
\end{equation}
where $w$ is the guidance strength.

\subsection{Architecture}
\label{subsec:architecture}

\begin{figure}[h]
  \centering
  \includegraphics[width=\textwidth]{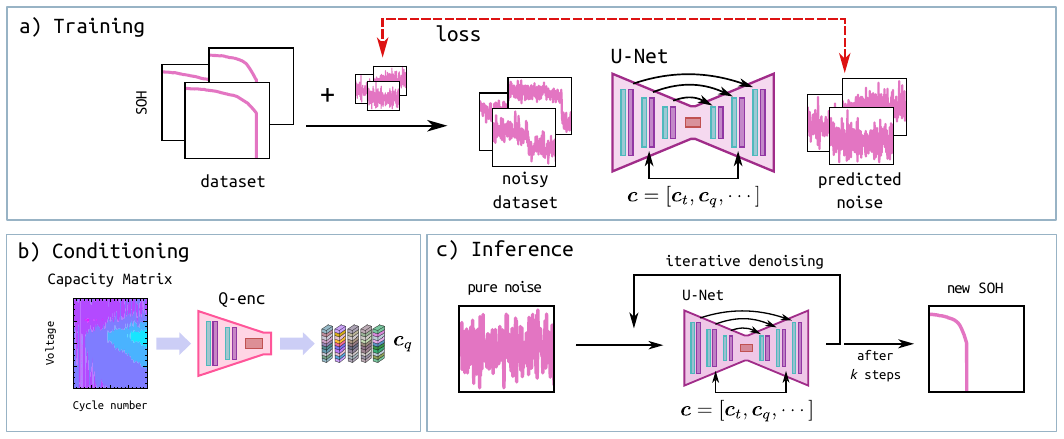}
  \caption{Schematic view of the model architecture. Adapted and modified from the work by \citet{Frrutter2024}, with permission from the authors. Modifications include context-specific changes.}
  \label{fig:architecture}
\end{figure}

Similar to the current state-of-the-art architectures for image and audio diffusion models \cite{DiffVision,yang2024surveydiffusionmodelstime}, DiffBatt is based on a U-Net architecture (see \cref{fig:architecture}\textcolor{red}{a}) and employs diffusion processes to generate SOH curves similar to a time series generation task. Conditioning for battery information, e.g., the capacity matrix, or a diffusion timestep $t$, is provided by adding embeddings into intermediate layers of the network \cite{ddpm}. The model consists of residual blocks with one-dimensional convolutions, attention modules, and pooling and up-sampling layers. In the reverse process, the model takes a one-dimensional Gaussian noise as input and generates a sample of SOH degradation. Notably, SOH typically exhibits a decreasing trend with progressive cycling. To better capture this physical behavior, we append a positional encoding to the output of the first convolution, allowing the denoising process to incorporate knowledge of the cycle number.~\Cref{fig:architecture} depicts a schematic view of the model architecture.

\begin{figure}[h]
  \centering
  \includegraphics[width=\textwidth]{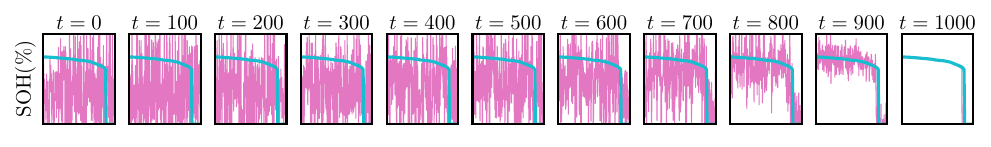}
  \caption{Denoising steps for one test sample of the \texttt{MATR} dataset.}
  \label{fig:record}
\end{figure}

DiffBatt can transform Gaussian noise into a new SOH curve through the reverse diffusion process, as illustrated in \cref{fig:architecture}\textcolor{red}{c} and \cref{fig:record}. For this study, we employ the concept of the \textit{capacity matrix} ($\mQ$), as introduced by \citet{attia2021statistical}, as an additional condition for the diffusion process. The capacity matrix serves as a compact representation of battery electrochemical cycling data, incorporating a series of feature representations. Consistent with prior research on machine learning for predicting battery degradation \cite{attia2021statistical,Severson2019Data-drivenDegradation,Zhang2023BatteryML:AnDegradation}, we utilize the capacity matrix corresponding to the first 100 cycles. This choice is driven by the high costs, time, and effort associated with long-term battery testing. Our goal is to leverage early life performance data to predict battery degradation and minimize resource expenditure. To encode $\mQ$ into an embedding ($\vc_q$), we utilize a transformer encoder (see \cref{fig:architecture}\textcolor{red}{b}). This allows DiffBatt to generate SOH curves, from which the RUL can be derived by calculating the number of cycles until the SOH drops below a specified threshold, such as 80\% of the nominal capacity.

\subsection{Data}

We conduct a comprehensive evaluation based on several publicly accessible battery datasets, as detailed in BatteryML \cite{Zhang2023BatteryML:AnDegradation}. The datasets included are \texttt{CALCE}~~\citep{CALCE2,CALCE1}, \texttt{HUST}~\citep{HUST}, \texttt{MATR}~\citep{NE2,Severson2019Data-drivenDegradation}, \texttt{RWTH}~\citep{RWTH}, \texttt{SNL}~\citep{SNL}, and \texttt{UL\_PUR}~\citep{UL_PUR,UL_PUR2}. These datasets encompass various battery chemistries including lithium iron phosphate (LFP), lithium cobalt oxide (LCO), nickel manganese cobalt oxide (NMC), nickel cobalt aluminum oxide (NCA), and a combination of NMC and LCO (NMC\_LCO). The datasets provide diverse information on battery materials, capacities, voltages, temperatures, and state of charge (SOC) and RUL ranges. We combine several datasets as recommended in the BatteryML \cite{Zhang2023BatteryML:AnDegradation} study: \texttt{CRUH} (combining \texttt{CALCE}, \texttt{RWTH}, \texttt{UL\_PUR}, and \texttt{HNEI}), \texttt{CRUSH} (combining \texttt{CALCE}, \texttt{RWTH}, \texttt{UL\_PUR}, \texttt{SNL}, and \texttt{HNEI}), and \texttt{MIX} (including all datasets except \texttt{SNL}). In addition, we utilize the data splits provided by BatteryML to maintain consistency and comparability, and we benchmark our results against those reported in their study. We refer to \cref{app:data} and \citet{Zhang2023BatteryML:AnDegradation} for more detailed information on the data.

\section{Results and discussion}
\label{sec:results}

Our model can be utilized for RUL prediction, SOH estimation, and SOH synthesis. It should be noted that we refer to generating new SOH curves for data augmentation as SOH synthesis. SOH and RUL prediction are two common tasks in managing inevitable capacity fade, widely discussed in the literature. While both tasks aim to predict capacity fade, they utilize different data representations. In this section, we conduct experiments based on the BatteryML benchmark tests \cite{Zhang2023BatteryML:AnDegradation} and compare the results. Furthermore, we explore the potential of our model to generate synthetic data, thereby augmenting limited battery degradation datasets for ML tasks. Further detailed analysis and results are included in \cref{app:soh,app:syn}.

\subsection{RUL prediction} 
We train DiffBatt with $p_{\rm uncond}$ of 0.2 on different datasets and compare it with the models from BatteryML \cite{Zhang2023BatteryML:AnDegradation}. For the prediction tasks, we generate ten SOH samples from ten different input noises for each sample of the capacity matrix. Then, we select the sample that best fits the first 100 cycles as the final prediction. The RUL is further computed from the predicted SOH curve. 

The results for the RUL prediction task are summarized in \cref{tab:rul-results}. This table illustrates the performance of the models in the RUL task using root-mean-squared error (RMSE) as the evaluation metric. RMSE is suitable for the RUL task since it represents the error based on an average number of cycles in which the predicted RUL differs from the reference. We train the DiffBatt model with ten different initialization seeds for each test and report the mean error along with the standard deviation, indicated as a subscript. DiffBatt shows notable performance improvements on multiple datasets. Specifically, DiffBatt achieved the lowest RMSE on the \texttt{MATR1}, \texttt{SNL}, and \texttt{CRUSH} datasets, with RMSE values of $88\pm4$, $125\pm11$, and $294\pm18$ respectively. These results highlight DiffBatt's robustness and precision in predicting RUL across different battery compositions and operational conditions. For instance, in the \texttt{SNL} dataset, DiffBatt's RMSE of $125\pm11$ outperforms the best benchmark model, PCR, which has an RMSE of 200 and is significantly lower than that of the CNN model, $924\pm267$, indicating a substantial improvement in predictive accuracy.

Furthermore, the comparative analysis shows that DiffBatt consistently performs better than other advanced models across various datasets. On the \texttt{MIX} dataset, DiffBatt achieved a mean RMSE of $202\pm6$, which is slightly higher than the best-performing model's RMSE of 197 but significantly lower than those of the deep learning models. Although DiffBatt did not achieve the lowest RMSE on the \texttt{MATR2} dataset ($235\pm16$), it remains competitive compared to other deep learning models. Importantly, DiffBatt exhibits a mean RMSE of 196 across all datasets, outperforming all other models and demonstrating superior generalizability. These results illustrate DiffBatt's efficacy in learning and generalizing from diverse data sources. By utilizing the data splits and benchmarking against results from BatteryML \cite{Zhang2023BatteryML:AnDegradation}, we ensure that our comparisons are both fair and indicative of DiffBatt's capabilities.

\begin{table}[h]
  \centering
  \begin{threeparttable}  
  \caption{Results obtained from DiffBatt for RUL prediction against benchmark results reported in \cite{Zhang2023BatteryML:AnDegradation}. For models sensitive to initialization, values in the table correspond to the mean error and the standard deviation, mean\textsubscript{std}, across ten seeds. Values that are \underline{underlined} represent the best results from the benchmarks in BatteryML, while values that are \textbf{bold} indicate the overall best results.}
  \label{tab:rul-results}
  \small
  \begin{tabular}{@{}lccccccccc@{}}
  \toprule
  \textbf{Models}     & \texttt{MATR1}  & \texttt{MATR2}   & \texttt{HUST}    & \texttt{SNL}  & \texttt{CLO}  & \texttt{CRUH} & \texttt{CRUSH} & \texttt{MIX} & Mean \\
  \midrule
  "Variance" model   & 136 & 211  & 398  & 360 & 179  & 118 & 506 & 521 & 304 \\
  "Discharge" model  & 329 & \underline{\textbf{149}} & \underline{\textbf{322}} & 267 & 143 & 76  & $>$1000 & $>$1000 & 411 \\
  "Full" model      & 167 & $>$1000 & 335 & 433 & \underline{\textbf{138}} & 93 & $>$1000 & 331 & 437 \\
  \midrule
  Ridge regression    & 116 & 184 & $>$1000 & 242 & 169 & 65 & $>$1000 & 372 & 268 \\
  PCR              &\underline{90} & 187 & 435 & \underline{200} &197 &68 & 560 &376 & 264 \\
  PLSR            &104 &181 & 431 &242 &176 &\underline{\textbf{60}} &535 &383 & 264 \\
  Gaussian process   &154 &224 &$>$1000 &251 &204 &115 &$>$1000 &573 & 440 \\
  XGBoost & 334 & 799 & 395 & 547 & 215 & 119 & \underline{330} & 205 & 368 \\
  Random forest      &168\textsubscript{9} &233\textsubscript{7} &368\textsubscript{7} &532\textsubscript{25} &192\textsubscript{2} &81\textsubscript{1} &416\textsubscript{5} &\underline{\textbf{197}\textsubscript{0}} & 273\\
  \midrule
  MLP    &149\textsubscript{3} &275\textsubscript{27} &459\textsubscript{9} &370\textsubscript{81} &146\textsubscript{5} &103\textsubscript{4} &565\textsubscript{9} &451\textsubscript{42} & \underline{263}\\
  CNN     &102\textsubscript{94} &228\textsubscript{104} &465\textsubscript{75} &924\textsubscript{267} &$>$1000 &174\textsubscript{92} &545\textsubscript{11} &272\textsubscript{101} & 464\\
  LSTM       &119\textsubscript{11} &219\textsubscript{33} &443\textsubscript{29} &539\textsubscript{40} &222\textsubscript{12} &105\textsubscript{10} &519\textsubscript{39} &268\textsubscript{9} & 304 \\
  Transformer    & 135\textsubscript{13} & 364\textsubscript{25} & 391\textsubscript{11} & 424\textsubscript{23} & 187\textsubscript{14} & 81\textsubscript{8} & 550\textsubscript{21} & 271\textsubscript{16} & 300 \\
  \midrule

    DiffBatt (ours) & \textbf{88}\textsubscript{4} & 235\textsubscript{16} & 368\textsubscript{23} & \textbf{125}\textsubscript{11} & 140\textsubscript{14} & 119\textsubscript{6} & \textbf{294}\textsubscript{18} & 202\textsubscript{6} & \textbf{196} \\
  \bottomrule
  \end{tabular}
  \end{threeparttable}  
\end{table}

\begin{figure}[h]
  \centering
  \begin{subfigure}[t]{0.30\textwidth}
   \centering
    \includegraphics[width=\textwidth]{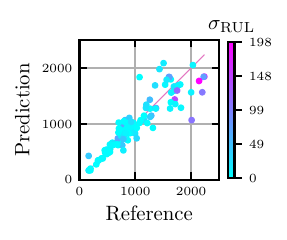}
    \caption{RUL prediction}
  \end{subfigure}
  \hfill
  \begin{subfigure}[t]{0.69\textwidth}
   \centering
      \includegraphics[width=\textwidth]{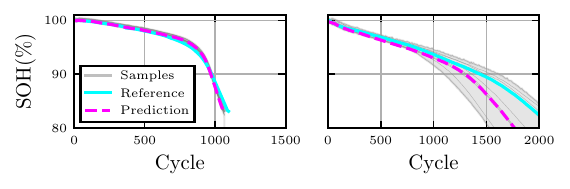}
      \caption{SOH prediction}
  \end{subfigure}
  \caption{Results obtained from DiffBatt for RUL prediction and SOH estimation on the \texttt{MIX} test datasets. (a) Predicted RUL against the reference, colored by the standard deviation $\sigma_{\rm RUL}$. (b) Generated samples and selected predictions based on the best fit to the first 100 cycles, compared against the reference SOH for test samples with the lowest (left) and highest (right) uncertainty in the predictions.}
  \label{fig:rul}
\end{figure}

In \cref{fig:rul}, we present the results for the RUL task for each test sample from the \texttt{MIX} dataset, obtained using the DiffBatt model. Additionally, the figure includes SOH curves that correspond to the predictions with the lowest and highest uncertainty. DiffBatt is capable of quantifying the uncertainty in its predictions, which is reported here as the standard deviation of the RUL, computed from ten generated samples. The data reveals that samples with higher prediction errors generally tend to exhibit larger deviations in RUL.

\subsection{SOH estimation}
In practical applications, estimating a battery's SOH requires predicting the current discharge capacity under standardized conditions using reference performance tests (RPTs) and historical cycling data. However, the discrepancy between real-world battery usage and these standardized conditions poses significant challenges in obtaining precise ground-truth labels, complicating accurate SOH estimation. Consequently, developing a robust benchmark test for SOH estimation remains an ongoing effort within the research community. Therefore, we benchmark our SOH estimation results by approximating the SOH under actual operating conditions. Results are reported in \cref{tab:soh-results}. We employ the same datasets in our RUL prediction tasks for the SOH estimation experiments, which allows for reproducibility due to the clear data splits. For each degradation curve, we calculate the error as the RMSE between the reference and predicted curves up to an SOH representing the end of life (EOL). The reference SOH is zero-padded to match the length of the predicted curve if necessary. Results are reported as the mean RMSE across test samples for different EOLs. To ensure reproducibility, a detailed discussion on experimental setup and error metrics is included in \cref{app:soh}.

DiffBatt demonstrates strong performance in SOH estimation, achieving low RMSE values across various datasets. When considering an EOL of 80\%, our model exhibits the highest precision with the \texttt{CRUH} dataset, yielding the lowest RMSE of $1.26\pm0.04$, and maintains consistent accuracy with RMSE values of $1.68\pm0.08$ and $1.89\pm0.03$ on the \texttt{MATR1} and \texttt{SNL} datasets, respectively. DiffBatt has a higher RMSE with the \texttt{HUST} dataset at $2.75\pm0.15$, and results in an RMSE of $2.17\pm0.07$ with the \texttt{MATR2} dataset. The \texttt{MIX} dataset, aggregating diverse data sources, results in an RMSE of $1.98\pm0.03$, indicating the model's ability to generalize effectively. Consistent RMSE values across the \texttt{CLO} ($2.32\pm0.07$) and \texttt{CRUSH} ($2.35\pm0.05$) datasets further highlight DiffBatt's robustness across different battery chemistries and operating conditions. To account for the varying EOL requirements across different battery applications, such as electric vehicles and storage systems, we also evaluated DiffBatt's performance for EOL values of 90\%, 70\%, and 60\%. At an EOL of 90\%, DiffBatt maintains strong precision with low RMSE values across all datasets. Despite the increased difficulty, DiffBatt demonstrates robust performance with a minor increase in RMSE values across different EOL thresholds, resulting in mean RMSE values of 1.19, 2.05, 2.59, and 3.16 for EOLs of 90\%, 80\%, 70\%, and 60\%, respectively. These results underscore DiffBatt's adaptability for reliable SOH estimations tailored to specific battery application requirements.

\begin{table}[h]
  \centering
  \begin{threeparttable}  
  \caption{Results obtained from DiffBatt for SOH estimation task considering different end of life (EOL) values. Reported results correspond to the mean RMSE and the standard deviation, mean\textsubscript{std}, across ten seeds.}
  \label{tab:soh-results}
  \small
  \begin{tabular}{@{}lccccccccc@{}}
  \toprule
  EOL & \texttt{MATR1}  & \texttt{MATR2}   & \texttt{HUST}    & \texttt{SNL}  & \texttt{CLO}  & \texttt{CRUH} & \texttt{CRUSH} & \texttt{MIX} & Mean \\
  \midrule
    90\% & $ 1.00_{\,0.07} $ & $ 1.36_{\,0.07} $ & $ 1.46_{\,0.1} $ & $ 0.98_{\,0.05} $ & $ 1.47_{\,0.06} $ & $ 0.68_{\,0.02} $ & $ 1.39_{\,0.03} $ & $ 1.21_{\,0.03} $ & $1.19$\\
    80\% & $ 1.68_{\,0.08} $ & $ 2.17_{\,0.07} $ & $ 2.75_{\,0.15} $ & $ 1.89_{\,0.03} $ & $ 2.32_{\,0.07} $ & $ 1.26_{\,0.04} $ & $ 2.35_{\,0.05} $ & $ 1.98_{\,0.03} $ & $2.05$\\
    70\% & $ 1.86_{\,0.08} $ & $ 2.83_{\,0.07} $ & $ 3.13_{\,0.16} $ & $ 2.40_{\,0.03} $ & $ 2.77_{\,0.07} $ & $ 2.00_{\,0.06} $ & $ 3.25_{\,0.09} $ & $ 2.51_{\,0.03} $ & $2.59$\\
    60\% & $ 2.23_{\,0.08} $ & $ 3.97_{\,0.07} $ & $ 3.43_{\,0.16} $ & $ 2.95_{\,0.04} $ & $ 3.16_{\,0.09} $ & $ 2.43_{\,0.07} $ & $ 3.97_{\,0.12} $ & $ 3.13_{\,0.02} $  & $3.16$\\
  \bottomrule
  \end{tabular}
  \end{threeparttable}  
\end{table}

\subsection{Battery degradation synthesis}

Conducting comprehensive battery degradation experiments in a real-world setting is often time-consuming and resource-intensive, requiring years to complete \cite{everlasting}. Synthesizing battery degradation data offers significant advantages by reducing the time and cost of long-term experiments and ensuring consistent, high-quality data. Our DiffBatt model can generate high-quality battery degradation curves, enabling the creation of large and diverse datasets that can enhance model training and robustness. To evaluate the effectiveness of our approach, we utilize the DiffBatt model trained on the \texttt{MATR1} dataset to generate 10 synthetic SOH curves for each sample in the training set, using varying rates for classifier-free diffusion guidance. We then leverage these synthesized SOH curves to train a random forest (RF) model for predicting RUL, while introducing 1\% Gaussian noise into the input capacity matrix for data augmentation. Each model is trained with ten different initialization seeds. To assess the quality and diversity of our generated samples, we use FID \cite{fid}, which provides insight into the similarity between real and synthetic images, Recall \cite{recall} as the main metric for measuring diversity and the RMSE of the random forest model on the test dataset, serving as a measure of its predictive accuracy. We also report Precision \cite{recall} for completeness. Note that FID is computed using principal component analysis (PCA) to map the data into a latent space instead of the Inception model. Results are summarized in \cref{tab:synthesis} against those reported in \cite{Zhang2023BatteryML:AnDegradation} for a random forest model trained without synthetic data. The synthetic data generated with varying guidance strengths are depicted in \cref{fig:syn} in the appendix.

On the one hand, the consistently low range of FID scores indicates the high quality of the synthetic data. The configuration with no guidance, $w=0$, achieves the lowest FID score of 0.405, indicating that this setting produces synthetic data most similar to the real data. However, as $w$ increases, the FID score rises, reaching 0.413 at $w=6.0$, indicating a slight decrease in the similarity between synthetic and real data.

On the other hand, Recall represents the diversity by measuring the fraction of the training data manifold covered by the generated samples. The highest Recall is obtained with $w=0.0$ and, as expected, sampling with a high guidance strength reduces diversity, leading to a Recall of 0.398 for $w=6.0$. This reduction suggests that while higher guidance strength might improve certain aspects of the synthetic data, it compromises the overall diversity.

The RMSE values provide insight into the predictive performance of the random forest model. The lowest RMSE of 104 is achieved with $w=2.0$, although the FID score is not the lowest in this case. This indicates that while the synthetic data generated at higher $w$ values diverges more from the real data (as per FID), it still offers a good balance between quality and diversity to enhance the model's predictive accuracy. Conversely, the RMSE for the random forest model trained on the real data reported in \cite{Zhang2023BatteryML:AnDegradation} was 168, significantly higher than the RMSE values observed for models trained on synthetic data.

\begin{table}[h]
  \centering
  \begin{threeparttable}  
  \caption{Results for the \texttt{MATR1} dataset obtained using random forests trained on synthetic data generated with different diffusion guidance $w$, against results reported in \cite{Zhang2023BatteryML:AnDegradation} obtained from a random forest trained without synthetic data.
  }
  \label{tab:synthesis}
  \small
  \begin{tabular}{@{}l|c|ccccc@{}}
  \toprule
  \multicolumn{4}{r}{} & \large{$w$} \\
  \cline{3-7}\\[-1.5ex]
  \textbf{Models} & RF \cite{Zhang2023BatteryML:AnDegradation} &
  $0.0$ &
  $1.0$ &
  $2.0$ &
  $4.0$ &
  $6.0$ \\
  \midrule
  FID ($\downarrow$) & NA & \textbf{0.405} & 0.408 & 0.409 & 0.411 & 0.413 \\
  Precision ($\uparrow$) & NA & \textbf{0.998} & 0.99 & 0.985 & 0.963 & 0.968 \\
  Recall ($\uparrow$) & NA & \textbf{0.663} & 0.663 & 0.614 & 0.446 & 0.398 \\
  RMSE  & 168\textsubscript{9} &
  109\textsubscript{6} &
  107\textsubscript{6} &
  \textbf{104}\textsubscript{6} &
  105\textsubscript{7} &
  106\textsubscript{8}  \\
\bottomrule
\end{tabular}
\end{threeparttable}  
\end{table} 

\section{Towards a foundation model for battery degradation}
\label{sec:foundation_model}

A foundation model is a large-scale, general-purpose artificial intelligence model pre-trained on extensive datasets, designed to be fine-tuned for various specific tasks with minimal additional training \cite{bommasani2022opportunitiesrisksfoundationmodels}. By training on several battery datasets and demonstrating strong generalizability and robustness across various tasks, DiffBatt offers a promising pathway toward creating such a foundation model for battery degradation. Inspired by the principles outlined in \cite{bommasani2022opportunitiesrisksfoundationmodels}, we explore how DiffBatt can be developed into a comprehensive foundation model for battery degradation.

\paragraph{Expressivity.}
DiffBatt is trained on the largest publicly available battery degradation datasets, encompassing a wide range of battery chemistries and operational conditions. This comprehensive training equips DiffBatt with a robust understanding of battery degradation patterns, making it an ideal candidate for foundational modeling. Our results demonstrate DiffBatt's high \textit{expressivity} and its ability to capture the complex dynamics involved in battery degradation. 

\paragraph{Multimodality.}
In DiffBatt, the input conditions are encoded using a transformer encoder that can be fine-tuned with relatively small amounts of new data from different cell chemistries. This ensures the model's quick adaptation to new battery technologies with minimal retraining. Furthermore, additional data modalities such as temperature, current profiles, and environmental conditions can be encoded and added to the condition vector, enhancing \textit{multimodality}. This multimodal approach ensures all relevant factors influencing battery health are considered, improving the model's accuracy and adaptability. 

\paragraph{Scalability and memory.}
DiffBatt's flexible architecture, leveraging diffusion models and transformers, ensures efficient \textit{scalability} and \textit{memory} usage. Its design allows it to handle extensive volumes of data and complex battery degradation scenarios effectively, making it capable of scaling with increasing data inputs without compromising performance. Additionally, DiffBatt's ability to synthesize battery degradation curves furthers its versatility by enabling the creation of large, diverse datasets. By generating synthetic degradation data, DiffBatt can augment existing datasets from cell chemistries with limited samples, enhancing the generalization capabilities of downstream models. 

\paragraph{Compositionality.}
DiffBatt exemplifies compositionality through its probabilistic model structure, which integrates conditional/unconditional diffusion models with transformer encoders. This architectural composition allows DiffBatt to leverage the strengths of both approaches, enabling flexible adaptation to a wide range of battery degradation tasks. The diffusion models handle probabilistic data generation, while the transformer encoders effectively capture and encode various input conditions. This compositional framework facilitates the transfer of learned parameters to new models, ensuring the efficient application of foundational knowledge to novel contexts and tasks.

By leveraging pre-training on diverse datasets and advanced architectural frameworks, DiffBatt provides a versatile and adaptable solution, capable of addressing a wide range of battery health prediction tasks. Our model can facilitate more reliable, efficient, and scalable battery technology advancements, ultimately contributing to the development of longer-lasting and more efficient energy storage solutions.

\section{Conclusions and outlook}

Tackling battery degradation is a major hurdle in advancing green technologies and sustainable energy solutions. Accurately predicting battery capacity loss remains particularly challenging due to its intricate and complex nature. To address this issue, we present DiffBatt, a novel general-purpose model for predicting and synthesizing battery degradation patterns based on diffusion models with classifier-free guidance and transformer encoders. A key innovation is the integration of conditional and unconditional diffusion models, enabling the robust generation of high-quality degradation curves. DiffBatt functions as both a probabilistic model to capture the inherent uncertainties in aging processes and a generative model to simulate and predict battery degradation over time.

We evaluate the performance of DiffBatt across three different tasks, i.e., RUL prediction, SOH estimation, and SOH synthesis. In the RUL prediction task, DiffBatt achieved the lowest RMSE on the \texttt{MATR1}, \texttt{SNL}, and \texttt{CRUSH} datasets, with RMSE values of $88\pm4$, $125\pm11$, and $294\pm18$ respectively. Notably, DiffBatt results in a mean RMSE of 196 across all datasets, significantly outperforming all other competing models. These results illustrate DiffBatt's efficacy in learning from and generalizing across diverse data sources. Consistently low RMSE values in the SOH prediction task further highlight DiffBatt's robustness and reliability across different battery chemistries. Moreover, we showcase the broad applicability of DiffBatt to generate high-quality battery degradation curves. 
We show that augmenting battery datasets with synthetic data can lead to a better and more accurate performance of downstream ML models, e.g., for RUL prediction.

We believe that by training on several diverse battery datasets and demonstrating strong generalizability and robustness across various tasks, DiffBatt offers a promising pathway toward developing a foundational model for battery degradation. However, to support a deep understanding of degradation mechanisms and derive counter measures in battery design and or battery production as well as formation the data-driven DiffBatt can be linked to physical-based models or needs to be extended with regard to the variation of battery design and process parameters.

\section*{Acknowledgement}
This research was conducted within the Research Training Group CircularLIB, supported by the Ministry of Science and Culture of Lower Saxony with funds from the program zukunft.niedersachsen of the Volkswagen Foundation (MWK | ZN3678). We would like to thank the anonymous reviewers who contributed to the improvement of our manuscript with their valuable suggestions and comments. We acknowledge that \cref{fig:architecture} is adapted and modified from the work by \citet{Frrutter2024}, available on arXiv \cite{Frrutter2024arXiv} under a CC BY-SA 4.0 license. The modifications include context-specific changes.

\setlength{\bibsep}{5pt}
\bibliographystyle{abbrvnat}
{\small \bibliography{references,main}}

\begin{thebibliography}{47}
\providecommand{\natexlab}[1]{#1}
\providecommand{\url}[1]{\texttt{#1}}
\expandafter\ifx\csname urlstyle\endcsname\relax
  \providecommand{\doi}[1]{doi: #1}\else
  \providecommand{\doi}{doi: \begingroup \urlstyle{rm}\Url}\fi

\bibitem[Attia et~al.(2021)Attia, Severson, and Witmer]{attia2021statistical}
P.~M. Attia, K.~A. Severson, and J.~D. Witmer.
\newblock Statistical learning for accurate and interpretable battery lifetime
  prediction.
\newblock \emph{Journal of The Electrochemical Society}, 168\penalty0
  (9):\penalty0 090547, 2021.
\newblock \doi{10.1149/1945-7111/ac2704}.
\newblock URL \url{https://dx.doi.org/10.1149/1945-7111/ac2704}.

\bibitem[Bastek et~al.(2024)Bastek, Sun, and
  Kochmann]{bastek2024physicsinformeddiffusionmodels}
J.-H. Bastek, W.~Sun, and D.~M. Kochmann.
\newblock Physics-informed diffusion models.
\newblock \emph{arXiv preprint arXiv:2403.14404}, 2024.
\newblock URL \url{https://arxiv.org/abs/2403.14404}.

\bibitem[Bl{\"{o}}meke et~al.(2022)Bl{\"{o}}meke, Scheller, Cerdas, Thies,
  Hachenberger, Gonter, Herrmann, and Spengler]{Blomeke2022}
S.~Bl{\"{o}}meke, C.~Scheller, F.~Cerdas, C.~Thies, R.~Hachenberger, M.~Gonter,
  C.~Herrmann, and T.~S. Spengler.
\newblock {Material and energy flow analysis for environmental and economic
  impact assessment of industrial recycling routes for lithium-ion traction
  batteries}.
\newblock \emph{Journal of Cleaner Production}, 377:\penalty0 134344, 12 2022.
\newblock ISSN 09596526.
\newblock \doi{10.1016/j.jclepro.2022.134344}.
\newblock URL
  \url{https://linkinghub.elsevier.com/retrieve/pii/S0959652622039166}.

\bibitem[Blomgren(2017)]{Blomgren2017TheBatteries}
G.~E. Blomgren.
\newblock The development and future of lithium ion batteries.
\newblock \emph{Journal of The Electrochemical Society}, 164\penalty0
  (1):\penalty0 A5019--A5025, 12 2017.
\newblock ISSN 0013-4651.
\newblock \doi{10.1149/2.0251701jes}.
\newblock URL \url{https://iopscience.iop.org/article/10.1149/2.0251701jes}.

\bibitem[Bommasani et~al.(2022)Bommasani, Hudson, Adeli, Altman, Arora, von
  Arx, Bernstein, Bohg, Bosselut, Brunskill, Brynjolfsson, Buch, Card,
  Castellon, Chatterji, Chen, Creel, Davis, Demszky, Donahue, Doumbouya,
  Durmus, Ermon, Etchemendy, Ethayarajh, Fei-Fei, Finn, Gale, Gillespie, Goel,
  Goodman, Grossman, Guha, Hashimoto, Henderson, Hewitt, Ho, Hong, Hsu, Huang,
  Icard, Jain, Jurafsky, Kalluri, Karamcheti, Keeling, Khani, Khattab, Koh,
  Krass, Krishna, Kuditipudi, Kumar, Ladhak, Lee, Lee, Leskovec, Levent, Li,
  Li, Ma, Malik, Manning, Mirchandani, Mitchell, Munyikwa, Nair, Narayan,
  Narayanan, Newman, Nie, Niebles, Nilforoshan, Nyarko, Ogut, Orr,
  Papadimitriou, Park, Piech, Portelance, Potts, Raghunathan, Reich, Ren, Rong,
  Roohani, Ruiz, Ryan, Ré, Sadigh, Sagawa, Santhanam, Shih, Srinivasan,
  Tamkin, Taori, Thomas, Tramèr, Wang, Wang, Wu, Wu, Wu, Xie, Yasunaga, You,
  Zaharia, Zhang, Zhang, Zhang, Zhang, Zheng, Zhou, and
  Liang]{bommasani2022opportunitiesrisksfoundationmodels}
R.~Bommasani, D.~A. Hudson, E.~Adeli, R.~Altman, S.~Arora, S.~von Arx, M.~S.
  Bernstein, J.~Bohg, A.~Bosselut, E.~Brunskill, E.~Brynjolfsson, S.~Buch,
  D.~Card, R.~Castellon, N.~Chatterji, A.~Chen, K.~Creel, J.~Q. Davis,
  D.~Demszky, C.~Donahue, M.~Doumbouya, E.~Durmus, S.~Ermon, J.~Etchemendy,
  K.~Ethayarajh, L.~Fei-Fei, C.~Finn, T.~Gale, L.~Gillespie, K.~Goel,
  N.~Goodman, S.~Grossman, N.~Guha, T.~Hashimoto, P.~Henderson, J.~Hewitt,
  D.~E. Ho, J.~Hong, K.~Hsu, J.~Huang, T.~Icard, S.~Jain, D.~Jurafsky,
  P.~Kalluri, S.~Karamcheti, G.~Keeling, F.~Khani, O.~Khattab, P.~W. Koh,
  M.~Krass, R.~Krishna, R.~Kuditipudi, A.~Kumar, F.~Ladhak, M.~Lee, T.~Lee,
  J.~Leskovec, I.~Levent, X.~L. Li, X.~Li, T.~Ma, A.~Malik, C.~D. Manning,
  S.~Mirchandani, E.~Mitchell, Z.~Munyikwa, S.~Nair, A.~Narayan, D.~Narayanan,
  B.~Newman, A.~Nie, J.~C. Niebles, H.~Nilforoshan, J.~Nyarko, G.~Ogut, L.~Orr,
  I.~Papadimitriou, J.~S. Park, C.~Piech, E.~Portelance, C.~Potts,
  A.~Raghunathan, R.~Reich, H.~Ren, F.~Rong, Y.~Roohani, C.~Ruiz, J.~Ryan,
  C.~Ré, D.~Sadigh, S.~Sagawa, K.~Santhanam, A.~Shih, K.~Srinivasan,
  A.~Tamkin, R.~Taori, A.~W. Thomas, F.~Tramèr, R.~E. Wang, W.~Wang, B.~Wu,
  J.~Wu, Y.~Wu, S.~M. Xie, M.~Yasunaga, J.~You, M.~Zaharia, M.~Zhang, T.~Zhang,
  X.~Zhang, Y.~Zhang, L.~Zheng, K.~Zhou, and P.~Liang.
\newblock On the opportunities and risks of foundation models.
\newblock \emph{arXiv preprint arXiv:2108.07258}, 2022.
\newblock URL \url{https://arxiv.org/abs/2108.07258}.

\bibitem[Brock et~al.(2019)Brock, Donahue, and
  Simonyan]{brock2019largescalegantraining}
A.~Brock, J.~Donahue, and K.~Simonyan.
\newblock Large scale {GAN} training for high fidelity natural image synthesis.
\newblock \emph{arXiv preprint arXiv:1809.11096}, 2019.
\newblock URL \url{https://arxiv.org/abs/1809.11096}.

\bibitem[Broussely et~al.(2005)Broussely, Biensan, Bonhomme, Blanchard,
  Herreyre, Nechev, and Staniewicz]{Broussely2005MainBatteries}
M.~Broussely, P.~Biensan, F.~Bonhomme, P.~Blanchard, S.~Herreyre, K.~Nechev,
  and R.~Staniewicz.
\newblock {Main aging mechanisms in Li-ion batteries}.
\newblock \emph{Journal of Power Sources}, 146\penalty0 (1-2):\penalty0 90--96,
  8 2005.
\newblock ISSN 03787753.
\newblock \doi{10.1016/j.jpowsour.2005.03.172}.
\newblock URL
  \url{https://linkinghub.elsevier.com/retrieve/pii/S0378775305005082}.

\bibitem[Chombo and Laoonual(2020)]{Chombo2020ABattery}
P.~V. Chombo and Y.~Laoonual.
\newblock {A review of safety strategies of a Li-ion battery}.
\newblock \emph{Journal of Power Sources}, 478:\penalty0 228649, 12 2020.
\newblock ISSN 03787753.
\newblock \doi{10.1016/j.jpowsour.2020.228649}.
\newblock URL
  \url{https://linkinghub.elsevier.com/retrieve/pii/S0378775320309538}.

\bibitem[Croitoru et~al.(2023)Croitoru, Hondru, Ionescu, and Shah]{DiffVision}
F.-A. Croitoru, V.~Hondru, R.~T. Ionescu, and M.~Shah.
\newblock Diffusion models in vision: {A} survey.
\newblock \emph{IEEE Transactions on Pattern Analysis and Machine
  Intelligence}, 45\penalty0 (9):\penalty0 10850--10869, 2023.
\newblock \doi{10.1109/TPAMI.2023.3261988}.
\newblock URL \url{https://ieeexplore.ieee.org/document/10081412}.

\bibitem[Cui et~al.(2022)Cui, Wang, Li, and Wang]{Cui2022ANetwork}
Z.~Cui, L.~Wang, Q.~Li, and K.~Wang.
\newblock {A comprehensive review on the state of charge estimation for
  lithium‐ion battery based on neural network}.
\newblock \emph{International Journal of Energy Research}, 46\penalty0
  (5):\penalty0 5423--5440, 4 2022.
\newblock ISSN 0363-907X.
\newblock \doi{10.1002/er.7545}.
\newblock URL \url{https://onlinelibrary.wiley.com/doi/10.1002/er.7545}.

\bibitem[Devie et~al.(2018)Devie, Baure, and Dubarry]{HNEI}
A.~Devie, G.~Baure, and M.~Dubarry.
\newblock Intrinsic variability in the degradation of a batch of commercial
  18650 lithium-ion cells.
\newblock \emph{Energies}, 11\penalty0 (5), 2018.
\newblock ISSN 1996-1073.
\newblock \doi{10.3390/en11051031}.
\newblock URL \url{https://www.mdpi.com/1996-1073/11/5/1031}.

\bibitem[Dhariwal and Nichol(2021)]{dhariwal2021diffusion}
P.~Dhariwal and A.~Q. Nichol.
\newblock Diffusion models beat {GAN}s on image synthesis.
\newblock In \emph{Advances in Neural Information Processing Systems}, 2021.
\newblock URL \url{https://openreview.net/forum?id=AAWuCvzaVt}.

\bibitem[Edge et~al.(2021)Edge, O’Kane, Prosser, Kirkaldy, Patel, Hales,
  Ghosh, Ai, Chen, Yang, Li, Pang, Bravo~Diaz, Tomaszewska, Marzook,
  Radhakrishnan, Wang, Patel, Wu, and Offer]{Edge2021LithiumKnow}
J.~S. Edge, S.~O’Kane, R.~Prosser, N.~D. Kirkaldy, A.~N. Patel, A.~Hales,
  A.~Ghosh, W.~Ai, J.~Chen, J.~Yang, S.~Li, M.-C. Pang, L.~Bravo~Diaz,
  A.~Tomaszewska, M.~W. Marzook, K.~N. Radhakrishnan, H.~Wang, Y.~Patel, B.~Wu,
  and G.~J. Offer.
\newblock {Lithium ion battery degradation: what you need to know}.
\newblock \emph{Physical Chemistry Chemical Physics}, 23\penalty0
  (14):\penalty0 8200--8221, 4 2021.
\newblock ISSN 1463-9076.
\newblock \doi{10.1039/D1CP00359C}.
\newblock URL \url{https://xlink.rsc.org/?DOI=D1CP00359C}.

\bibitem[Fürrutter et~al.(2024{\natexlab{a}})Fürrutter, Muñoz-Gil, and
  Briegel]{Frrutter2024}
F.~Fürrutter, G.~Muñoz-Gil, and H.~J. Briegel.
\newblock Quantum circuit synthesis with diffusion models.
\newblock \emph{Nature Machine Intelligence}, 6\penalty0 (5):\penalty0
  515--524, May 2024{\natexlab{a}}.
\newblock ISSN 2522-5839.
\newblock \doi{10.1038/s42256-024-00831-9}.
\newblock URL \url{http://dx.doi.org/10.1038/s42256-024-00831-9}.

\bibitem[Fürrutter et~al.(2024{\natexlab{b}})Fürrutter, Muñoz-Gil, and
  Briegel]{Frrutter2024arXiv}
F.~Fürrutter, G.~Muñoz-Gil, and H.~J. Briegel.
\newblock Quantum circuit synthesis with diffusion models.
\newblock \emph{arXiv preprint arXiv:2311.02041}, 2024{\natexlab{b}}.
\newblock URL \url{https://arxiv.org/abs/2311.02041}.

\bibitem[Galeotti et~al.(2015)Galeotti, Cin{\`{a}}, Giammanco, Cordiner, and
  Di~Carlo]{Galeotti2015PerformanceSpectroscopy}
M.~Galeotti, L.~Cin{\`{a}}, C.~Giammanco, S.~Cordiner, and A.~Di~Carlo.
\newblock {Performance analysis and SOH (state of health) evaluation of lithium
  polymer batteries through electrochemical impedance spectroscopy}.
\newblock \emph{Energy}, 89:\penalty0 678--686, 9 2015.
\newblock ISSN 03605442.
\newblock \doi{10.1016/j.energy.2015.05.148}.
\newblock URL
  \url{https://linkinghub.elsevier.com/retrieve/pii/S0360544215007756}.

\bibitem[Ginster et~al.(2024)Ginster, Bl{\"{o}}meke, Popien, Scheller, Cerdas,
  Herrmann, and Spengler]{Ginster2024CircularImpacts}
R.~Ginster, S.~Bl{\"{o}}meke, J.~Popien, C.~Scheller, F.~Cerdas, C.~Herrmann,
  and T.~S. Spengler.
\newblock {Circular battery production in the EU: Insights from integrating
  life cycle assessment into system dynamics modeling on recycled content and
  environmental impacts}.
\newblock \emph{Journal of Industrial Ecology}, 28\penalty0 (5):\penalty0
  1165--1182, 10 2024.
\newblock ISSN 1088-1980.
\newblock \doi{10.1111/jiec.13527}.
\newblock URL \url{https://onlinelibrary.wiley.com/doi/10.1111/jiec.13527}.

\bibitem[He et~al.(2011)He, Williard, Osterman, and Pecht]{CALCE2}
W.~He, N.~Williard, M.~Osterman, and M.~Pecht.
\newblock Prognostics of lithium-ion batteries based on {Dempster--Shafer}
  theory and the {Bayesian Monte Carlo method}.
\newblock \emph{Journal of Power Sources}, 196\penalty0 (23):\penalty0
  10314--10321, 2011.
\newblock \doi{https://doi.org/10.1016/j.jpowsour.2011.08.040}.
\newblock URL
  \url{https://www.sciencedirect.com/science/article/pii/S0378775311015400}.

\bibitem[Heusel et~al.(2017)Heusel, Ramsauer, Unterthiner, Nessler, and
  Hochreiter]{fid}
M.~Heusel, H.~Ramsauer, T.~Unterthiner, B.~Nessler, and S.~Hochreiter.
\newblock {GANs} trained by a two time-scale update rule converge to a local
  {Nash} equilibrium.
\newblock In \emph{Advances in Neural Information Processing Systems},
  volume~30, 2017.
\newblock URL
  \url{https://proceedings.neurips.cc/paper_files/paper/2017/file/8a1d694707eb0fefe65871369074926d-Paper.pdf}.

\bibitem[Ho and Salimans(2022)]{diffusionguidance}
J.~Ho and T.~Salimans.
\newblock Classifier-free diffusion guidance.
\newblock \emph{arXiv preprint arXiv:2207.12598}, 2022.
\newblock URL \url{https://arxiv.org/abs/2207.12598}.

\bibitem[Ho et~al.(2020)Ho, Jain, and Abbeel]{ddpm}
J.~Ho, A.~Jain, and P.~Abbeel.
\newblock Denoising diffusion probabilistic models.
\newblock In \emph{Advances in Neural Information Processing Systems},
  volume~33, 2020.
\newblock URL
  \url{https://proceedings.neurips.cc/paper_files/paper/2020/file/4c5bcfec8584af0d967f1ab10179ca4b-Paper.pdf}.

\bibitem[Ho et~al.(2022)Ho, Saharia, Chan, Fleet, Norouzi, and
  Salimans]{Ho2022}
J.~Ho, C.~Saharia, W.~Chan, D.~J. Fleet, M.~Norouzi, and T.~Salimans.
\newblock Cascaded diffusion models for high fidelity image generation.
\newblock \emph{Journal of Machine Learning Research}, 23\penalty0
  (47):\penalty0 1--33, 2022.
\newblock URL \url{http://jmlr.org/papers/v23/21-0635.html}.

\bibitem[Hong et~al.(2020)Hong, Lee, Jeong, and Yi]{NE2}
J.~Hong, D.~Lee, E.-R. Jeong, and Y.~Yi.
\newblock Towards the swift prediction of the remaining useful life of
  lithium-ion batteries with end-to-end deep learning.
\newblock \emph{Applied Energy}, 278:\penalty0 115646, 2020.
\newblock ISSN 0306-2619.
\newblock \doi{https://doi.org/10.1016/j.apenergy.2020.115646}.
\newblock URL
  \url{https://www.sciencedirect.com/science/article/pii/S0306261920311429}.

\bibitem[Juarez-Robles et~al.(2020)Juarez-Robles, Jeevarajan, and
  Mukherjee]{UL_PUR}
D.~Juarez-Robles, J.~A. Jeevarajan, and P.~P. Mukherjee.
\newblock Degradation-safety analytics in lithium-ion cells: {Part I}. aging
  under charge/discharge cycling.
\newblock \emph{Journal of The Electrochemical Society}, 167\penalty0
  (16):\penalty0 160510, nov 2020.
\newblock \doi{10.1149/1945-7111/abc8c0}.
\newblock URL \url{https://dx.doi.org/10.1149/1945-7111/abc8c0}.

\bibitem[Juarez-Robles et~al.(2021)Juarez-Robles, Azam, Jeevarajan, and
  Mukherjee]{UL_PUR2}
D.~Juarez-Robles, S.~Azam, J.~A. Jeevarajan, and P.~P. Mukherjee.
\newblock Degradation-safety analytics in lithium-ion cells and modules: {Part
  III}. aging and safety of pouch format cells.
\newblock \emph{Journal of The Electrochemical Society}, 168\penalty0
  (11):\penalty0 110501, nov 2021.
\newblock \doi{10.1149/1945-7111/ac30af}.
\newblock URL \url{https://dx.doi.org/10.1149/1945-7111/ac30af}.

\bibitem[Kingma and Dhariwal(2018)]{Glow}
D.~P. Kingma and P.~Dhariwal.
\newblock Glow: {Generative} flow with invertible 1x1 convolutions.
\newblock In \emph{Advances in Neural Information Processing Systems},
  volume~31, 2018.
\newblock URL
  \url{https://proceedings.neurips.cc/paper_files/paper/2018/file/d139db6a236200b21cc7f752979132d0-Paper.pdf}.

\bibitem[Kynk\"{a}\"{a}nniemi et~al.(2019)Kynk\"{a}\"{a}nniemi, Karras, Laine,
  Lehtinen, and Aila]{recall}
T.~Kynk\"{a}\"{a}nniemi, T.~Karras, S.~Laine, J.~Lehtinen, and T.~Aila.
\newblock Improved precision and recall metric for assessing generative models.
\newblock In \emph{Advances in Neural Information Processing Systems},
  volume~32, 2019.
\newblock URL
  \url{https://proceedings.neurips.cc/paper/2019/hash/0234c510bc6d908b28c70ff313743079-Abstract.html}.

\bibitem[Li et~al.(2022)Li, West, and Preindl]{Li2022TowardsReview}
A.~G. Li, A.~C. West, and M.~Preindl.
\newblock {Towards unified machine learning characterization of lithium-ion
  battery degradation across multiple levels: A critical review}.
\newblock \emph{Applied Energy}, 316:\penalty0 119030, 6 2022.
\newblock ISSN 03062619.
\newblock \doi{10.1016/j.apenergy.2022.119030}.
\newblock URL
  \url{https://linkinghub.elsevier.com/retrieve/pii/S0306261922004354}.

\bibitem[Li et~al.(2024)Li, Biferale, Bonaccorso, Scarpolini, and
  Buzzicotti]{Li2024}
T.~Li, L.~Biferale, F.~Bonaccorso, M.~A. Scarpolini, and M.~Buzzicotti.
\newblock Synthetic {Lagrangian} turbulence by generative diffusion models.
\newblock \emph{Nature Machine Intelligence}, 6\penalty0 (4):\penalty0
  393--403, Apr 2024.
\newblock ISSN 2522-5839.
\newblock \doi{10.1038/s42256-024-00810-0}.
\newblock URL \url{https://doi.org/10.1038/s42256-024-00810-0}.

\bibitem[Li et~al.(2021)Li, Sengupta, Dechent, Howey, Annaswamy, and
  Sauer]{RWTH}
W.~Li, N.~Sengupta, P.~A. Dechent, D.~Howey, A.~Annaswamy, and D.~U. Sauer.
\newblock {O}ne-shot battery degradation trajectory prediction with deep
  learning.
\newblock \emph{Journal of Power Sources}, page 230024, 2021.
\newblock ISSN 0378-7753.
\newblock \doi{10.1016/j.jpowsour.2021.230024}.
\newblock URL \url{https://publications.rwth-aachen.de/record/820366}.

\bibitem[Li et~al.(2023)Li, Yu, S{\o}ren~Byg, and
  Daniel~Ioan]{Li2023TheBatteries}
X.~Li, D.~Yu, V.~S{\o}ren~Byg, and S.~Daniel~Ioan.
\newblock {The development of machine learning-based remaining useful life
  prediction for lithium-ion batteries}.
\newblock \emph{Journal of Energy Chemistry}, 82:\penalty0 103--121, 7 2023.
\newblock ISSN 20954956.
\newblock \doi{10.1016/j.jechem.2023.03.026}.
\newblock URL
  \url{https://linkinghub.elsevier.com/retrieve/pii/S2095495623001870}.

\bibitem[Luo et~al.(2023)Luo, Zhang, Zhu, and
  Li]{Luo2023State-of-HealthLearning}
C.~Luo, Z.~Zhang, S.~Zhu, and Y.~Li.
\newblock State-of-health prediction of lithium-ion batteries based on
  diffusion model with transfer learning.
\newblock \emph{Energies}, 16\penalty0 (9):\penalty0 3815, 4 2023.
\newblock ISSN 1996-1073.
\newblock \doi{10.3390/en16093815}.
\newblock URL \url{https://www.mdpi.com/1996-1073/16/9/3815}.

\bibitem[Luzi et~al.(2024)Luzi, Mayer, Casco-Rodriguez, Siahkoohi, and
  Baraniuk]{luzi2024boomeranglocalsamplingimage}
L.~Luzi, P.~M. Mayer, J.~Casco-Rodriguez, A.~Siahkoohi, and R.~G. Baraniuk.
\newblock Boomerang: {Local} sampling on image manifolds using diffusion
  models.
\newblock \emph{arXiv preprint arXiv:2210.12100}, 2024.
\newblock URL \url{https://arxiv.org/abs/2210.12100}.

\bibitem[Ma et~al.(2022)Ma, Xu, Jiang, Cheng, Yang, Shen, Yang, Huang, Ding,
  and Yuan]{HUST}
G.~Ma, S.~Xu, B.~Jiang, C.~Cheng, X.~Yang, Y.~Shen, T.~Yang, Y.~Huang, H.~Ding,
  and Y.~Yuan.
\newblock Real-time personalized health status prediction of lithium-ion
  batteries using deep transfer learning.
\newblock \emph{Energy \& Environmental Science}, 2022.
\newblock \doi{10.1039/D2EE01676A}.
\newblock URL \url{http://dx.doi.org/10.1039/D2EE01676A}.

\bibitem[O'Kane et~al.(2022)O'Kane, Ai, Madabattula, Alonso-Alvarez, Timms,
  Sulzer, Edge, Wu, Offer, and Marinescu]{OKane2022Lithium-ionIt}
S.~E.~J. O'Kane, W.~Ai, G.~Madabattula, D.~Alonso-Alvarez, R.~Timms, V.~Sulzer,
  J.~S. Edge, B.~Wu, G.~J. Offer, and M.~Marinescu.
\newblock {Lithium-ion battery degradation: how to model it}.
\newblock \emph{Physical Chemistry Chemical Physics}, 24\penalty0
  (13):\penalty0 7909--7922, 3 2022.
\newblock ISSN 1463-9076.
\newblock \doi{10.1039/D2CP00417H}.
\newblock URL \url{https://xlink.rsc.org/?DOI=D2CP00417H}.

\bibitem[Preger et~al.(2020)Preger, Barkholtz, Fresquez, Campbell, Juba,
  Romàn-Kustas, Ferreira, and Chalamala]{SNL}
Y.~Preger, H.~M. Barkholtz, A.~Fresquez, D.~L. Campbell, B.~W. Juba,
  J.~Romàn-Kustas, S.~R. Ferreira, and B.~Chalamala.
\newblock Degradation of commercial lithium-ion cells as a function of
  chemistry and cycling conditions.
\newblock \emph{Journal of The Electrochemical Society}, 167\penalty0
  (12):\penalty0 120532, sep 2020.
\newblock \doi{10.1149/1945-7111/abae37}.
\newblock URL \url{https://dx.doi.org/10.1149/1945-7111/abae37}.

\bibitem[Rauf et~al.(2022)Rauf, Khalid, and Arshad]{Rauf2022MachineModelling}
H.~Rauf, M.~Khalid, and N.~Arshad.
\newblock {Machine learning in state of health and remaining useful life
  estimation: Theoretical and technological development in battery degradation
  modelling}.
\newblock \emph{Renewable and Sustainable Energy Reviews}, 156:\penalty0
  111903, 3 2022.
\newblock ISSN 13640321.
\newblock \doi{10.1016/j.rser.2021.111903}.
\newblock URL
  \url{https://linkinghub.elsevier.com/retrieve/pii/S1364032121011692}.

\bibitem[Ren and Du(2023)]{Ren2023ABatteries}
Z.~Ren and C.~Du.
\newblock {A review of machine learning state-of-charge and state-of-health
  estimation algorithms for lithium-ion batteries}.
\newblock \emph{Energy Reports}, 9:\penalty0 2993--3021, 12 2023.
\newblock ISSN 23524847.
\newblock \doi{10.1016/j.egyr.2023.01.108}.
\newblock URL
  \url{https://linkinghub.elsevier.com/retrieve/pii/S235248472300118X}.

\bibitem[Rubenbauer and Henninger(2017)]{Rubenbauer2017DefinitionsGrids}
H.~Rubenbauer and S.~Henninger.
\newblock {Definitions and reference values for battery systems in electrical
  power grids}.
\newblock \emph{Journal of Energy Storage}, 12:\penalty0 87--107, 8 2017.
\newblock ISSN 2352152X.
\newblock \doi{10.1016/j.est.2017.04.004}.
\newblock URL
  \url{https://linkinghub.elsevier.com/retrieve/pii/S2352152X16301797}.

\bibitem[Severson et~al.(2019)Severson, Attia, Jin, Perkins, Jiang, Yang, Chen,
  Aykol, Herring, Fraggedakis, Bazant, Harris, Chueh, and
  Braatz]{Severson2019Data-drivenDegradation}
K.~A. Severson, P.~M. Attia, N.~Jin, N.~Perkins, B.~Jiang, Z.~Yang, M.~H. Chen,
  M.~Aykol, P.~K. Herring, D.~Fraggedakis, M.~Z. Bazant, S.~J. Harris, W.~C.
  Chueh, and R.~D. Braatz.
\newblock {Data-driven prediction of battery cycle life before capacity
  degradation}.
\newblock \emph{Nature Energy}, 4\penalty0 (5):\penalty0 383--391, 3 2019.
\newblock ISSN 2058-7546.
\newblock \doi{10.1038/s41560-019-0356-8}.
\newblock URL \url{https://www.nature.com/articles/s41560-019-0356-8}.

\bibitem[Vetter et~al.(2005)Vetter, Nov{\'{a}}k, Wagner, Veit, M{\"{o}}ller,
  Besenhard, Winter, Wohlfahrt-Mehrens, Vogler, and
  Hammouche]{Vetter2005AgeingBatteries}
J.~Vetter, P.~Nov{\'{a}}k, M.~Wagner, C.~Veit, K.-C. M{\"{o}}ller,
  J.~Besenhard, M.~Winter, M.~Wohlfahrt-Mehrens, C.~Vogler, and A.~Hammouche.
\newblock {Ageing mechanisms in lithium-ion batteries}.
\newblock \emph{Journal of Power Sources}, 147\penalty0 (1-2):\penalty0
  269--281, 9 2005.
\newblock ISSN 03787753.
\newblock \doi{10.1016/j.jpowsour.2005.01.006}.
\newblock URL
  \url{https://linkinghub.elsevier.com/retrieve/pii/S0378775305000832}.

\bibitem[voor Technologisch Onderzoek~(VITO) et~al.(2021)voor Technologisch
  Onderzoek~(VITO), \`a l'Energie Atomique et~aux Energies Alternatives~(CEA),
  Siemens, (TUM), Testing, ALGOLiON, University, Smart, Eindhoven, Voltia, and
  Solutions]{everlasting}
V.~I. voor Technologisch Onderzoek~(VITO), C.~\`a l'Energie Atomique et~aux
  Energies Alternatives~(CEA), Siemens, T.~U.~M. (TUM), T.~S.~B. Testing,
  ALGOLiON, R.~A. University, L.~Smart, T.~U. Eindhoven, Voltia, and V.~E.~T.
  Solutions.
\newblock Everlasting: Electric vehicle enhanced range, lifetime and safety
  through ingenious battery management, 2021.
\newblock URL
  \url{https://data.4tu.nl/collections/EVERLASTING_Electric_Vehicle_Enhanced_Range_Lifetime_And_Safety_Through_INGenious_battery_management_/5065445/11}.

\bibitem[Wang et~al.(2021)Wang, Jin, Bai, Fan, Shi, and Fernandez]{Wang2021}
S.~Wang, S.~Jin, D.~Bai, Y.~Fan, H.~Shi, and C.~Fernandez.
\newblock A critical review of improved deep learning methods for the remaining
  useful life prediction of lithium-ion batteries.
\newblock \emph{Energy Reports}, 7:\penalty0 5562--5574, 2021.
\newblock ISSN 23524847.
\newblock \doi{10.1016/j.egyr.2021.08.182}.
\newblock URL \url{files/1917/A critical review of improved deep learning
  methods for the remaining.pdf https://doi.org/10.1016/j.egyr.2021.08.182}.

\bibitem[Xing et~al.(2013)Xing, Ma, Tsui, and Pecht]{CALCE1}
Y.~Xing, E.~W. Ma, K.-L. Tsui, and M.~Pecht.
\newblock An ensemble model for predicting the remaining useful performance of
  lithium-ion batteries.
\newblock \emph{Microelectronics Reliability}, 53\penalty0 (6):\penalty0
  811--820, 2013.
\newblock \doi{https://doi.org/10.1016/j.microrel.2012.12.003}.
\newblock URL
  \url{https://www.sciencedirect.com/science/article/pii/S0026271412005227}.

\bibitem[Yang et~al.(2024)Yang, Jin, Wen, Zhang, Liang, Ma, Wang, Liu, Yang,
  Xu, Bian, Pan, and Wen]{yang2024surveydiffusionmodelstime}
Y.~Yang, M.~Jin, H.~Wen, C.~Zhang, Y.~Liang, L.~Ma, Y.~Wang, C.~Liu, B.~Yang,
  Z.~Xu, J.~Bian, S.~Pan, and Q.~Wen.
\newblock A survey on diffusion models for time series and spatio-temporal
  data.
\newblock \emph{arXiv preprint arXiv:2404.18886}, 2024.
\newblock URL \url{https://arxiv.org/abs/2404.18886}.

\bibitem[Zhang et~al.(2024{\natexlab{a}})Zhang, Xu, Chen, and
  Zhuang]{Zhang2024}
B.~Zhang, P.~Xu, X.~Chen, and Q.~Zhuang.
\newblock Generative quantum machine learning via denoising diffusion
  probabilistic models.
\newblock \emph{Physical Review Letters}, 132:\penalty0 100602, Mar
  2024{\natexlab{a}}.
\newblock \doi{10.1103/PhysRevLett.132.100602}.
\newblock URL \url{https://link.aps.org/doi/10.1103/PhysRevLett.132.100602}.

\bibitem[Zhang et~al.(2024{\natexlab{b}})Zhang, Gui, Zheng, Lu, Li, and
  Bian]{Zhang2023BatteryML:AnDegradation}
H.~Zhang, X.~Gui, S.~Zheng, Z.~Lu, Y.~Li, and J.~Bian.
\newblock Battery{ML}: An open-source platform for machine learning on battery
  degradation.
\newblock In \emph{The Twelfth International Conference on Learning
  Representations}, 2024{\natexlab{b}}.
\newblock URL \url{http://arxiv.org/abs/2310.14714}.

\end{thebibliography}

\newpage
\appendix

\section{Appendix}

\subsection{Datasets}
\label{app:data}

\begin{figure}[h]
  \centering
  \includegraphics[width=0.9\textwidth]{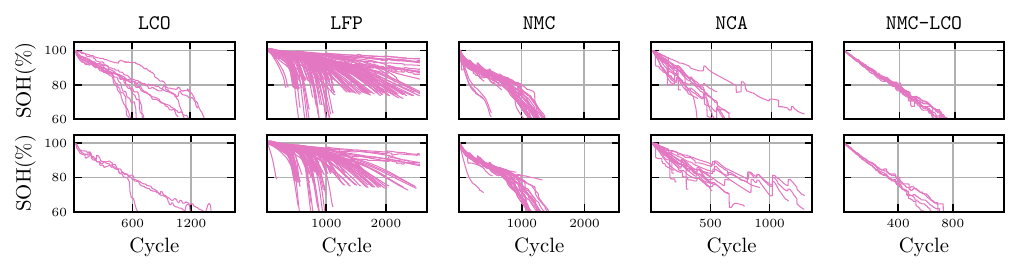}
  \caption{Train (up) and test (bottom) samples for each cell chemistry. The data is scaled using the SOH of the first cycle.}
  \label{fig:data}
\end{figure}

Battery degradation curves utilized for training and testing the models are depicted in \cref{fig:data} for each cell chemistry. A brief summary of the datasets included in BatteryML \cite{Zhang2023BatteryML:AnDegradation} is presented here.

The \texttt{CALCE} dataset includes full lifecycle data from 13 batteries with an LCO cathode. Each battery has a nominal capacity of 1100 mAh. They were all charged using a constant current/constant voltage protocol: 0.5C current until reaching 4.2V, maintaining 4.2V until the current dropped below 0.05A, and a cutoff voltage of 2.7V \citep{CALCE1, CALCE2}. 

The \texttt{MATR} dataset, provided by \citet{Severson2019Data-drivenDegradation} and \citet{NE2}, is one of the largest public datasets containing 180 commercial 18650 LFP batteries. These batteries, cycled at a forced convection temperature chamber of \SI{30}{\degree}C, have a nominal capacity of 1.1 Ah and a nominal voltage of 3.3V. The dataset comprises three subsets: \texttt{MATR1}, \texttt{MATR2} \cite{Severson2019Data-drivenDegradation}, and \texttt{CLO} \cite{NE2}, all categorized due to distinct measurement batches. 

The \texttt{HUST} dataset includes 77 LFP batteries, similar to those in the \texttt{MATR} dataset. These batteries followed an identical charging protocol with varying multi-stage discharge protocols, all conducted at a constant temperature of \SI{30}{\degree}C \citep{HUST}. 

The \texttt{HNEI} dataset contains 14 commercial 18650 cells with a graphite anode and a blended NMC and LCO cathode. These cells were cycled at 1.5C to 100\% depth of discharge for over 1000 cycles at room temperature \citep{HNEI}. 

The \texttt{SNL} dataset includes 61 commercial 18650 cells (NCA, NMC, and LFP), cycled to 80\% capacity. The study evaluates the impact of temperature, depth of discharge, and discharge current on long-term degradation \citep{SNL}. 

The \texttt{UL\_PUR} dataset comprises 10 commercial pouch cells with a graphite negative electrode and an NCA cathode. These cells were cycled at 1C between 2.7V and 4.2V, equivalent to 0-100\% state of charge (SOC), at room temperature until reaching 10-20\% capacity fade. Additionally, modules were cycled at C/2 between 13.7V and 21.0V until 20\% capacity fade \citep{UL_PUR, UL_PUR2}. 

The \texttt{RWTH} dataset contains data from 48 lithium-ion battery cells aged under identical conditions. These cells feature a carbon anode and an NMC cathode \citep{RWTH}. The cells were cycled at a constant ambient temperature of \SI{25}{\degree}C. Each cycle involved a 30-minute discharge phase down to 3.5V and a 30-minute charge phase up to 3.9V, with the currents capped at a maximum of 4A. This resulted in cycles between approximately 20\% and 80\% state of charge.

\subsection{SOH prediction}
\label{app:soh}

For prediction tasks, i.e. we employ a guidance strength of $w=0.0$ and generate ten samples for each input capacity matrix. The capacity matrix is constructed from the first 100 cycles. We further select the final prediction based on the best fit to the SOH of the first 100 cycles. The RMSE for an SOH sample $j$ is computed as
\begin{equation}
  {\rm RMSE}_j = \sqrt{\frac{1}{n_j}\sum_{i=1}^{n_j}({\tilde{y}}_i - {y}_i)^2}
\end{equation} 
where $\tilde{y}$ and $y$ represent the predicted and the reference SOH in percentage, respectively, $i$ denotes the cycle number and $n_j$ is the cycle number at which the predicted SOH reaches the EOL. Further, we report the mean RMSE across all the test samples as the RMSE for the dataset.

\begin{figure}[h]
  \centering
  \includegraphics[width=0.9\textwidth]{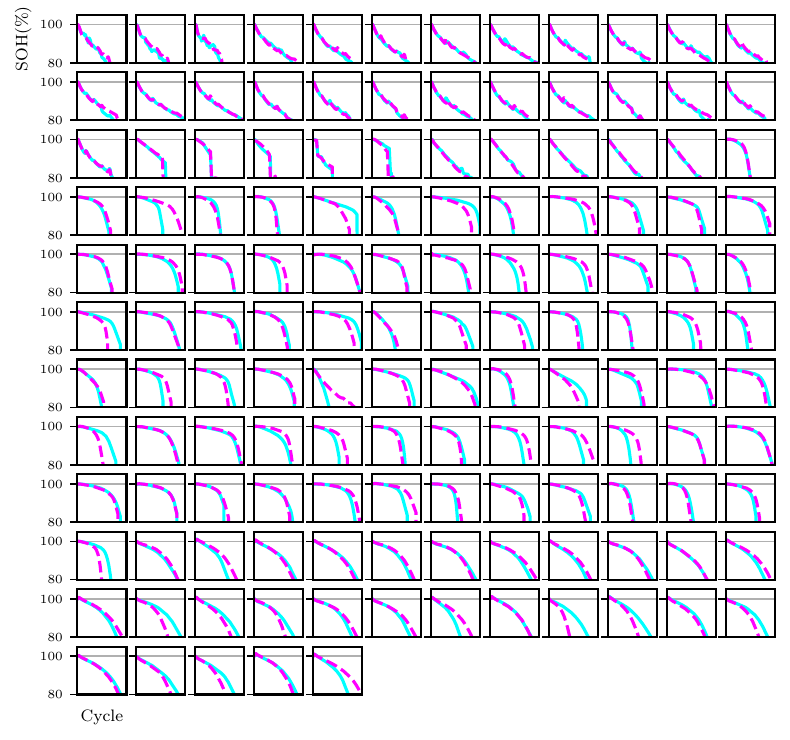}
  \caption{SOH predictions against reference for all the test samples of \texttt{MIX} dataset. The pink dashed line shows the prediction and the cyan solid line shows the reference.}
  \label{fig:mix}
\end{figure}

\Cref{fig:mix} illustrates the predicted SOH versus the reference SOH for all test samples in the \texttt{MIX} dataset. The results demonstrate that DiffBatt effectively captures various degradation dynamics and accurately predicts SOH for the majority of test samples and highlights DiffBatt's ability to generalize across different battery chemistries and operational conditions present in the \texttt{MIX} dataset. This capability is essential for developing reliable battery health monitoring systems that can adapt to diverse usage patterns and environmental factors. 

\subsection{SOH synthesis}
\label{app:syn}

\begin{figure}[th]
  \centering
  \includegraphics[width=0.9\textwidth]{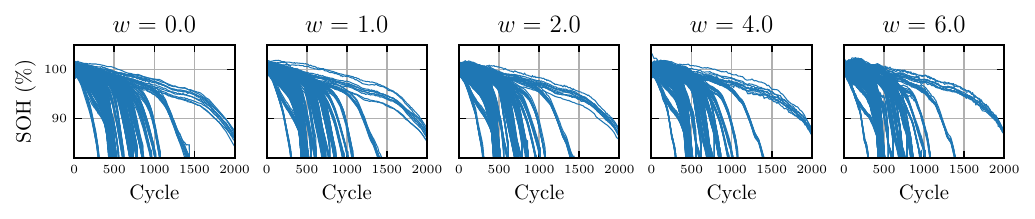}
  \caption{\texttt{MATR1} synthetic data generated by DiffBatt with different guidance strengths.}
  \label{fig:syn}
\end{figure}


\end{document}